\documentclass[lettersize,journal]{IEEEtran}
\usepackage{amsmath,amsfonts}
\usepackage{algorithmic}
\usepackage{algorithm}
\usepackage{array}
\usepackage[caption=false,font=normalsize,labelfont=sf,textfont=sf]{subfig}
\usepackage{textcomp}
\usepackage{stfloats}
\usepackage{url}
\usepackage{verbatim}
\usepackage{graphicx}
\usepackage{cite}
\usepackage{booktabs}
\usepackage[hidelinks]{hyperref}
\graphicspath{{figures/}{./}}
\newcommand{\tablefont}{\footnotesize}
\newcommand{\method}{EEG-HealthLLM}
\begin{document}
\title{EEG-HealthLLM: Hierarchical Adaptation of Large Language Models for Unified Heterogeneous EEG Health Assessment}
\author{Ruidong Ma, Jisheng Dang,   Ronghao Lin,   Bimei Wang, Jingwen Zhao, Wenjun Ma, Wei-Shi Zheng, 
 ~\IEEEmembership{Senior Member, ~IEEE}, Meng Wang, ~\IEEEmembership{Fellow, ~IEEE}, Qi Tian, ~\IEEEmembership{Fellow, ~IEEE}, Tat-Seng Chua
\thanks{This work was supported by the National Natural Science Foundation of China (Grants No. 62227807 and U24B20186). This work was also supported by the Supercomputing Center of Lanzhou University.}
\thanks{Ruidong Ma, Jisheng Dang,  Bimei Wang, Jingwen Zhao, and Wenjun Ma are with School of Information Science \& Engineering, Lanzhou University, Lanzhou 730000, China.  (E-mail: dangjsh@mail2.sysu.edu.cn).}

\thanks{Ronghao Lin is with the College of Computer Science and Software Engineering, Shenzhen University, Shenzhen 518060, China (E-mail: linrh7@mail2.sysu.edu.cn)}

\thanks{Wei-Shi Zheng is with the School of Computer Science and Engineering, Sun Yat-sen University, Guangzhou 510006, China (E-mail: wszheng@ieee.org).}

\thanks{Meng Wang is with School of Computer Science and Information Engineering, Hefei University of Technology, Hefei 230009, China. (E-mail: eric.mengwang@gmail.com).}

\thanks{Qi Tian is with Cloud and AI BU, Huawei, Shenzhen, Guangdong 518129,
China (e-mail: tian.qi1@huawei.com).}
\thanks{Tat-Seng Chua is with the School of Computing, National University of Singapore, Singapore 119077 (E-mail: dcscts@nus.edu.sg). }
\thanks{*Corresponding author:  Jisheng Dang and Ronghao Lin.}}

\maketitle

\begin{abstract}
    Electroencephalography (EEG) provides a non-invasive means for
    neurological and mental health assessment, yet existing EEG models are
    typically developed for individual tasks and are difficult to unify
    across heterogeneous signal configurations and assessment objectives.
    Large language models (LLMs) offer a common interface for multiple
    tasks, but adapting heterogeneous EEG signals to a shared LLM raises a
    key challenge: how to preserve task-specific signal characteristics
    while enabling effective language-side parameter sharing. To address
    this problem, we propose EEG-HealthLLM, a unified and
    parameter-efficient framework for heterogeneous EEG health assessment.
    We introduce an Asymmetric Hierarchical Adaptation Strategy:
    task-specific EEG adapters preserve heterogeneous spatio-temporal
    dynamics, while a unified shared LoRA is injected into the LLM backbone
    (e.g., Phi-3) to capture shared high-level assessment semantics across
    tasks. We further establish EEG-HealthBench by reorganizing five
    existing EEG resources into paradigms covering epileptic seizure
    detection, major depressive disorder assessment, anxiety risk
    assessment, depressive symptom risk assessment, and emotional state
    assessment under a unified subject-independent protocol. Extensive
    experiments show that EEG-HealthLLM achieves an average balanced
    accuracy (BAC) of 69.58\%, ranking first among the evaluated methods in
    both average BAC and macro-F1. Compared with independent single-task
    LoRA adaptation, it improves average BAC from 67.70\% to 69.58\% while
    reducing the aggregate language-side adapter footprint by 80.0\%.
    Further analyses reveal task-dependent cross-task interactions and
    trade-offs, supporting the proposed hierarchical adaptation strategy.
    The implementation is publicly available at
    \url{https://github.com/Marui-88/EEG-HealthLLM}.
    \end{abstract}


\section{Introduction}

Large language models (LLMs) have become an increasingly important
interface for medical information understanding, clinical reasoning,
and health assessment. Recent studies have examined the development,
clinical use, deployment, and practical limitations of medical LLMs,
highlighting both their potential and the challenges associated with
real-world medical applications
~\cite{thirunavukarasu2023llmmedicine,liu2025llmmedicine}.
Beyond task-specific systems, recent advances in generalist medical
artificial intelligence (AI) and generative AI have increasingly
emphasized reusable foundation models capable of supporting
heterogeneous medical modalities and diverse clinical tasks, together
with emerging capabilities such as agents, mixture-of-experts
architectures, and reasoning models
~\cite{moor2023generalist,teo2025generative}.

Compared with classifiers designed separately for each task, a
language interface can organize different prediction problems as
``medical evidence, task question, and structured answer''. This
formulation supports medical question answering, report generation,
and structured assessment. Recent vision-language studies show that
non-textual medical evidence can be connected to language models.
LLaVA advances visual instruction tuning~\cite{liu2023llava}, and
LLaVA-Med adapts this paradigm to biomedical image question answering
~\cite{li2023llavamed}. Med-Flamingo, PMC-VQA, RadFM, and BiomedGPT
further explore medical visual question answering and biomedical
vision-language modeling
~\cite{moor2023medflamingo,zhang2024pmcvqa,wu2023radfm,zhang2023biomedgpt}.
HealthGPT introduces heterogeneous low-rank adaptation (H-LoRA) and
staged training to unify medical visual comprehension and generation
~\cite{lin2025healthgpt}. These studies show that non-textual evidence
can be represented as continuous conditional tokens or cross-modal
embeddings. Nevertheless, most existing systems focus on medical
images, whereas continuous electroencephalography (EEG) signals remain
underexplored.

Beyond medical imaging, video understanding has explored complementary
approaches to representation learning and multimodal adaptation.
Decoupled representations support video segmentation and
grounding~\cite{dang2025decoupled,dang2026vdn}, while sparse propagation,
contextual memory, and selective memory retrieval support efficient
temporal modeling~\cite{dang2026fasttrack,dang2024beyond,dang2023imsfr,dang2025asm,dang2024tspsm}.
Preference optimization has also been investigated for detailed video
captioning~\cite{dang2026synpo}. These studies offer broader methodological
context for organizing modality-specific information and connecting it
with semantic representations. Adapting these principles to heterogeneous
EEG health assessment requires attention to differences in signal
characteristics, acquisition configurations, and task semantics.

EEG records brain electrical activity in a non-invasive manner and has
been widely used in neurological screening, mental health assessment,
and emotional-state analysis
~\cite{tveit2023scoreai,cai2022modma,xu2025emoeegmc}. 
More broadly, recent neurotechnology research has emphasized
translating brain-interfacing systems toward clinical use. Schalk et al.
reviewed engineering, scientific, and translational barriers to
clinical and commercial deployment~\cite{schalk2024translation}. 
For neuropsychiatric applications,
Oganesian and Shanechi reviewed brain--computer interface (BCI) approaches for decoding symptom
states and enabling personalized closed-loop interventions, highlighting
the importance of inter- and intra-individual neural variability
~\cite{oganesian2024bci}. In communication-oriented neuroprostheses,
Silva et al. summarized recent progress in decoding cortical activity
into text and speech, illustrating the growing clinical potential of
neural decoding~\cite{silva2024speech}. AI-assisted EEG analysis is
also becoming increasingly relevant to clinical neuroscience. Lucas
et al. reviewed AI applications in epilepsy across EEG, neuroimaging,
electronic health records, and other clinical modalities, covering
seizure detection and prediction as well as pathways toward clinical
translation~\cite{lucas2024epilepsy}. From a broader EEG perspective,
Zhang and Chen highlighted EEG connectomes as a framework for studying
brain function and dysfunction, and discussed the potential of machine
learning and adaptive closed-loop paradigms for cognitive and clinical
neuroscience~\cite{zhang2025eegconnectome}. Lee et al. further
demonstrated a non-invasive EEG-based BCI augmented with AI copilots
and shared autonomy for cursor and robotic-arm control, illustrating
how modern AI can complement neural decoding~\cite{lee2025aibci}.

Despite this progress, incorporating EEG into language-based medical
systems is fundamentally different from extending image-centered
multimodal models. EEG is a temporally dense and strongly
subject-dependent signal whose spatial organization, acquisition
protocol, and statistical properties can vary substantially across
datasets and assessment settings. Consequently, an EEG--LLM framework
must not only bridge the modality gap between neural signals and
language representations, but also preserve task- and dataset-specific
signal characteristics during cross-subject adaptation. Automated EEG
assessment is therefore challenging not only because of the complexity
of neural signals themselves, but also because heterogeneous tasks may
require different signal-processing pathways before their information
can be expressed within a common language space.

In EEG-based health assessment, different tasks address questions such
as whether abnormal EEG events are present, whether depression-related
patterns are detectable, whether anxiety or depressive symptom risk is
elevated, or which emotional state is reflected in an EEG recording.
If each task is modeled only as an isolated classifier, output formats
and task semantics remain separate. By connecting EEG signal to an LLM, each
problem can instead be represented as ``EEG evidence, a task-specific
assessment question, and a closed-set answer''. This formulation
provides a common output interface while keeping the answer space finite
and parseable.

The central challenge is to balance low-level signal heterogeneity with
high-level language-side sharing. EEG health assessment tasks differ in
acquisition protocol, channel configuration, sampling rate, temporal
window, and label semantics. Directly sharing a single EEG front-end
may mix inconsistent spatio-temporal patterns at the representation
stage. Conversely, training an independent EEG--LLM model for each task
requires separate language-side adaptation parameters and weakens the
value of a unified question-answering interface. The key question is
therefore how to preserve task-specific EEG alignment while sharing
language-side adaptation. More generally, multi-task learning studies
have shown that indiscriminate knowledge sharing across dissimilar tasks
can cause negative transfer, motivating more structured sharing across
tasks and features~\cite{yang2021taskfeature}. This setting therefore
calls for asymmetric sharing: specialization is required before
EEG--language alignment to preserve task-dependent signal structure,
whereas parameter sharing becomes more plausible after heterogeneous
EEG representations have been mapped into a common language embedding
space.

Existing EEG modeling and EEG--language studies have advanced
representation learning, heterogeneous biosignal modeling, and
EEG--language alignment, but they do not fully address this setting.
Discriminative EEG models such as EEGNet, DeepConvNet, ShallowConvNet,
EEG Conformer, and LMDA-Net learn spatio-temporal features from
multichannel EEG
~\cite{lawhern2018eegnet,schirrmeister2017deep,song2023eegconformer,miao2023lmda}.
However, they are usually trained for a specific task and label space.
BIOT extends representation learning to heterogeneous biosignals
~\cite{yang2023biot}, but is primarily designed as a transferable
biosignal representation model rather than a unified language-based
health assessment framework. EEG--language models such as NeuroLM and
Thought2Text further connect neural representations with language
models~\cite{jiang2024neurolm,mishra2025thought2text}, primarily
targeting EEG representation learning or brain-signal-to-text
generation rather than heterogeneous closed-set health assessment.
They are therefore not specifically designed for heterogeneous
cross-subject EEG health assessment under a unified medical language
interface.

To address this gap, we propose \method{}, a Phi-3-based adaptation
framework for heterogeneous cross-subject EEG health assessment. It
adapts the multimodal LLM architecture to heterogeneous EEG signals
through task-specific EEG pathways while retaining a shared Phi-3
language backbone and generative output interface. The framework uses
task-specific EEG adapters for signal-side alignment and shares a
language-side low-rank adaptation (LoRA) adapter across tasks. We further establish
EEG-HealthBench, a five-paradigm benchmark suite constructed from five
existing EEG resources under a unified subject-independent evaluation
protocol and a closed-set generative question answering (QA) format.
Experiments show that \method{} improves average balanced accuracy over
the single-task LoRA pipeline and achieves competitive task-averaged
performance against representative task-specific EEG classifiers.

The main contributions are summarized as follows:
\begin{itemize}
    \item We propose \method{}, a Phi-3-based adaptation framework for
    heterogeneous cross-subject EEG health assessment. It maps EEG
    signals into continuous EEG tokens through task-specific EEG
    adapters and organizes multiple EEG health assessment tasks as
    closed-set generative QA.

    \item We introduce a hierarchical modeling strategy that combines
    low-level task-specific EEG alignment with high-level shared
    language-side LoRA adaptation. This design preserves task-specific
    EEG structure and statistics, enables cross-task sharing of
    language-side representations, and reduces redundant language-side
    adapter storage across tasks.

    \item We establish EEG-HealthBench, a five-paradigm benchmark suite
    constructed from five existing EEG resources for heterogeneous EEG
    health assessment. It places epileptic seizure detection, major
    depressive disorder assessment, anxiety risk assessment, depressive
    symptom risk assessment, and emotional state assessment under a
    unified subject-independent protocol and closed-set generative QA
    format.
\end{itemize}

\section{Related Work}

\subsection{Medical Large Language Models}

Medical LLMs have been developed to improve medical knowledge understanding, medical question answering, and clinical text reasoning. HuatuoGPT uses medical question-answer pairs generated and refined from ChatGPT to train medical conversational models~\cite{zhang2023huatuogpt}. HuatuoGPT-II further unifies medical training corpora from different sources and formats into instruction-answer pairs to enhance medical reasoning~\cite{chen2023huatuogptii}. HuatuoGPT-o1 emphasizes medical reasoning through verifiable medical problems and reinforcement learning~\cite{chen2024huatuogpto1}. In parallel, medical multimodal models extend LLMs beyond text. LLaVA-Med, PMC-VQA, RadFM, BiomedGPT, Med-Flamingo, and HealthGPT connect medical images with language models through visual encoders, projectors, instruction tuning, or parameter-efficient adapters~\cite{li2023llavamed,zhang2024pmcvqa,wu2023radfm,zhang2023biomedgpt,moor2023medflamingo,lin2025healthgpt}. Beyond medical-domain models, BLIP-2 demonstrates that a frozen modality encoder and a frozen language model can be connected through a lightweight trainable bridge for efficient cross-modal adaptation~\cite{li2023blip2}. These models are important for medical image understanding, but their primary non-textual input remains visual evidence. At the same time, realistic clinical evaluations have exposed important limitations of current LLMs. Hager et al. showed that strong performance on medical examinations does not necessarily translate into reliable clinical decision-making, particularly with respect to information gathering, guideline adherence, laboratory interpretation, and workflow integration~\cite{hager2024clinical}. These findings motivate careful evaluation and appropriately constrained use of LLMs in medical applications. Adaptation of language backbones for continuous EEG health assessment remains less explored.

\subsection{EEG Health Assessment}

EEG health assessment has been studied in seizure screening, depression-related EEG analysis, anxiety risk assessment, depressive symptom risk assessment, and emotional state assessment. Early automated EEG analysis often relied on hand-crafted features and traditional classifiers. Spatial filtering methods such as common spatial pattern (CSP) and filter bank common spatial pattern (FBCSP) extract discriminative spatial or spectral-spatial patterns from multichannel EEG~\cite{ramoser2000csp,ang2008fbcsp}, but they usually depend on task-specific feature engineering and preprocessing. More recent deep EEG classifiers, including EEGNet, DeepConvNet, ShallowConvNet, EEG Conformer, LMDA-Net, and Dual-TSST, have shown that neural networks can learn useful spatial, temporal, and spectral patterns from multichannel EEG~\cite{lawhern2018eegnet,schirrmeister2017deep,song2023eegconformer,miao2023lmda,li2025dualtsst}. However, EEG signals are highly subject-dependent and are sensitive to recording protocol, channel montage, temporal window, and preprocessing pipeline. Cross-subject generalization is therefore a central challenge. More broadly, domain adaptation for multivariate time series has investigated invariant temporal mechanisms for mitigating distribution shifts across domains~\cite{cai2026timeseries}. Recent cross-subject emotion-recognition studies have addressed this problem through subject-invariant representation learning, class-confusion regularization, covariance alignment, and hyperbolic contrastive learning. Shen et al. learned subject-invariant EEG representations using contrastive objectives, whereas Cui et al. combined a gated recurrent unit with minimum class confusion for subject-independent emotion recognition~\cite{shen2023contrastive,cui2023subjectindependent}. Wang et al. introduced sparse Bayesian learning with enhanced covariance alignment, and Chang et al. proposed multi-scale hyperbolic contrastive learning for cross-subject emotion recognition~\cite{wang2025sbleca,chang2025mshcl}. 
LibEER further systematizes EEG emotion-recognition evaluation through
a comprehensive benchmark and algorithm library~\cite{liu2025libeer}.
Together, these methods and benchmarking efforts highlight the
importance of cross-subject robustness. However, the modeling
approaches considered above are primarily discriminative and optimized
for individual tasks with task-specific output heads; they do not
provide a unified generative language interface across heterogeneous
EEG health assessment tasks.

\subsection{EEG Foundation Models}

Recent EEG foundation models and biosignal representation models aim to
improve transferability across subjects, datasets, and downstream tasks.
BIOT learns token-based representations for heterogeneous biosignals
~\cite{yang2023biot}, while LaBraM, CBraMod, DMAE-EEG, EEG-DLite, and
EEG-DINO explore large-scale EEG pretraining, masked modeling,
self-distillation, and data-efficient pretraining strategies
~\cite{jiang2024labram,wang2025cbramod,zhang2025dmaeeeg,tang2026eegdlite,wang2025eegdino}.
These approaches improve the reuse of EEG or physiological
representations across downstream settings, but the resulting models
typically still rely on task-specific prediction heads and output
spaces. Thus, transferable signal representations alone do not provide
a common semantic interface for heterogeneous EEG health assessment
tasks with different acquisition protocols and label definitions.

Beyond EEG-specific models, task- and domain-transfer studies have shown
that structuring task-specific feature spaces can facilitate knowledge
transfer across tasks and generalization to unseen domains
~\cite{ramirez2023learning}. EEG-language studies further investigate
how neural signals can be connected with language models, and recent
surveys summarize this emerging direction~\cite{babu2025llmeegsurvey}.
NeuroLM and Thought2Text explore EEG tokenization, pretrained neural
representations, or brain-signal-to-text generation
~\cite{jiang2024neurolm,mishra2025thought2text}. These studies establish
important connections between neural representations and language
models, but their main objectives differ from closed-set heterogeneous
health assessment, where task semantics and valid answer spaces are
explicitly constrained.

IntervEEG-LLM further explores EEG-based multimodal data for customized
mental health interventions~\cite{zhang2025interveegllm}. However, the
setting considered in our work additionally requires multiple
heterogeneous EEG health assessment tasks to be handled under a common
language interface while preserving differences in signal distributions,
acquisition configurations, and assessment semantics. The challenge is
therefore not only to learn transferable EEG representations, but also
to determine which representations and adaptation parameters should
remain task specific and which can be shared after alignment to a
common language space.

In contrast, our work does not aim to build a universal EEG foundation
model or an open-ended brain-to-text generator. Instead, we focus on
LLM adaptation for heterogeneous EEG health assessment, where multiple
EEG tasks are organized as closed-set generative QA with task-specific
EEG adapters and shared language-side adaptation.
\subsection{Parameter-Efficient Fine-Tuning}

Adapting EEG to a language backbone for medical assessment requires both signal representation and efficient language-side adaptation. Full-parameter fine-tuning updates the entire backbone, but it is expensive and may overfit when medical data are limited or heterogeneous. Parameter-efficient fine-tuning freezes most backbone parameters and trains a small set of additional parameters. Representative approaches include adapter tuning~\cite{houlsby2019adapter}, prefix tuning~\cite{li2021prefixtuning}, prompt tuning~\cite{lester2021prompttuning}, and low-rank adaptation (LoRA)~\cite{hu2021lora}. Parameter-efficient adaptation has also been explored for pretrained multimodal networks, where lightweight trainable modules adapt frozen representations to changing modality conditions with limited additional parameter cost~\cite{reza2025robust}. In medical multimodal modeling, non-textual inputs are often mapped into continuous conditional representations, and lightweight language-side adapters are used for cross-modal alignment. HealthGPT introduces H-LoRA to decouple heterogeneous knowledge associated with medical visual comprehension and generation and employs routed LoRA experts within task-type-specific adapters~\cite{lin2025healthgpt}. In contrast, our framework shares a routed multi-branch LoRA adapter across heterogeneous EEG health assessment tasks. More broadly, multi-task architectures have explored selective parameter sharing through combinations of shared and task-specific representations, as well as task-specific attention over shared features~\cite{misra2016crossstitch,liu2019mtan}. For heterogeneous EEG health assessment, the key issue is not only how to adapt an LLM efficiently, but also which parameters should be shared across tasks. This motivates our hierarchical design: low-level EEG signal differences are handled by task-specific EEG adapters, whereas high-level closed-set QA adaptation is performed through a shared language-side LoRA adapter.

\section{Method}

\begin{figure*}[tbp]
    \centering
    \includegraphics[width=0.92\textwidth]{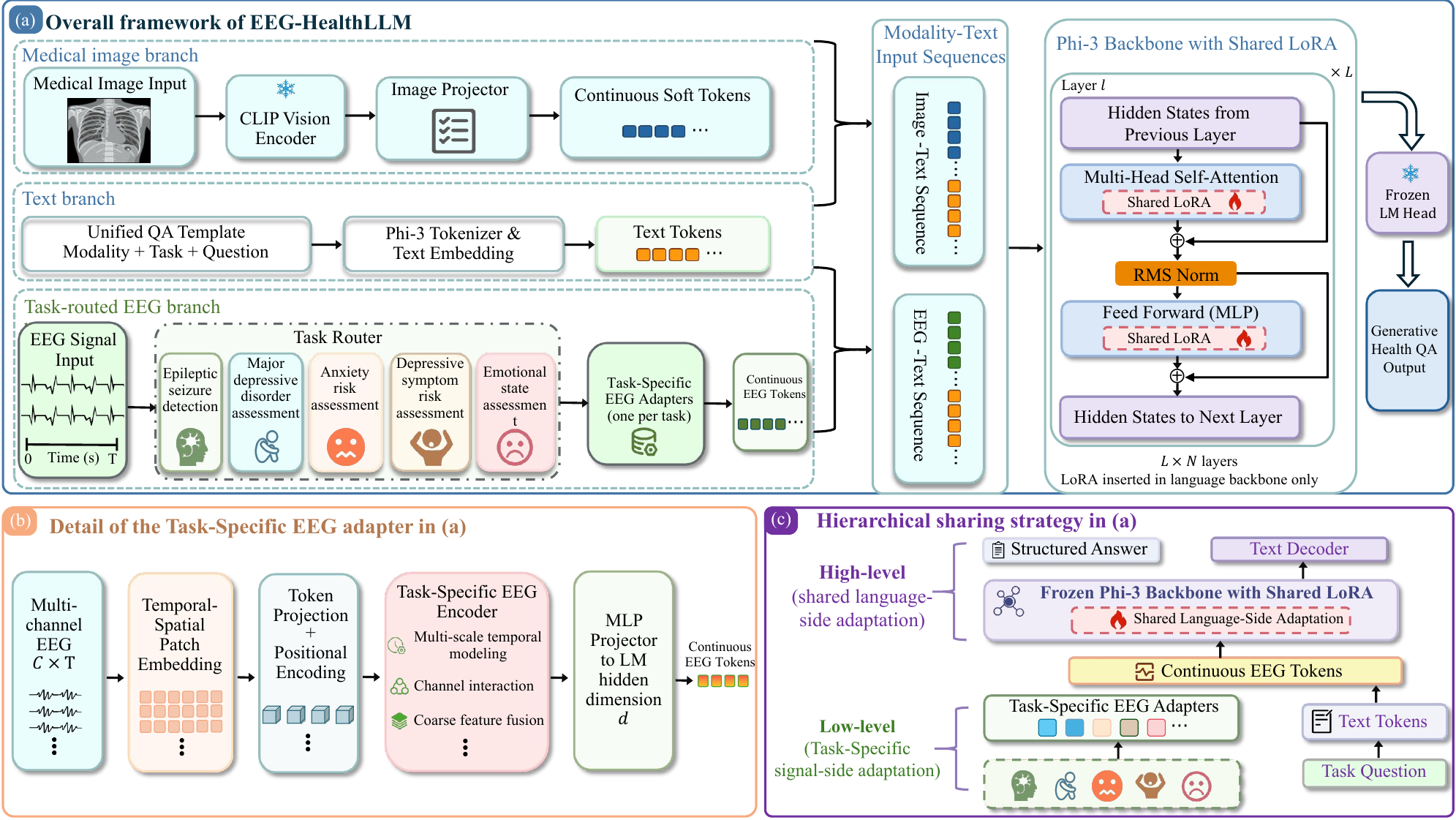}
    \caption{Overall framework of \method{}.
    (a) Overall architecture.
    (b) Task-specific EEG adapter for continuous EEG token generation.
    (c) Hierarchical sharing strategy with task-specific signal-side adaptation and shared language-side LoRA. The medical-image pathway is shown for architectural context, while the experiments focus on EEG.}
    \label{fig:framework}
\end{figure*}

\subsection{Problem Formulation}

Different EEG health assessment tasks correspond to different acquisition protocols, signal distributions, label spaces, and assessment objectives. If an independent classifier is built for each task, model outputs remain restricted to separate task-specific label spaces. 
We therefore formulate EEG health assessment as closed-set
generative QA, i.e., generative classification in which each
generated output is mapped to an answer in a predefined finite
answer set. Given an EEG signal and a task-specific assessment question, the model generates a parseable health assessment answer from that set. This formulation does not imply open-ended medical reasoning.

Let the set of EEG health assessment tasks be
\begin{equation}
\mathcal{R}=\{1,2,3,4,5\}.
\end{equation}
For the $r$-th task, the $i$-th EEG input is denoted by $x^{\mathrm{eeg}}_{r,i}$, the corresponding task-specific assessment question is $q_r$, and the target answer is $y_{r,i}$. Each task has a predefined answer set
\begin{equation}
\mathcal{A}_r=\{a_{r,1},a_{r,2},\ldots,a_{r,K_r}\},
\end{equation}
where $K_r$ is the number of candidate answers for task $r$, and $y_{r,i}\in \mathcal{A}_r$. These answer sets correspond to epileptic seizure detection, major depressive disorder assessment, anxiety risk assessment, depressive symptom risk assessment, and emotional state assessment.

We first map the EEG signal $x^{\mathrm{eeg}}_{r,i}$ into continuous EEG tokens, which serve as conditioning tokens for the frozen Phi-3 language backbone and are denoted by $z^{\mathrm{eeg}}_{r,i}$. The Phi-3 language backbone then takes these tokens and the task question as conditional inputs and autoregressively generates the target answer sequence. Given $y_{r,i}=(y_{r,i,1},\ldots,y_{r,i,T_i})$, the training objective is the conditional autoregressive cross-entropy loss:
{\small
\begin{equation}
\mathcal{L}_{\mathrm{QA}}
=
-
\sum_{r\in\mathcal{R}}
\frac{1}{N_r}
\sum_{i=1}^{N_r}
\frac{1}{T_i}
\sum_{t=1}^{T_i}
\log p_{\theta}
\bigl(
y_{r,i,t}
\mid
y_{r,i,<t}, q_r, z^{\mathrm{eeg}}_{r,i}
\bigr).
\label{eq:qa_loss}
\end{equation}
}
Here, $N_r$ denotes the effective number of training instances for task $r$ after task-specific balancing within an epoch, and is not necessarily the raw dataset size. $T_i$ is the length of the target answer sequence, $q_r$ specifies the current assessment semantics, and $z^{\mathrm{eeg}}_{r,i}$ provides EEG evidence related to the question. Equation~\eqref{eq:qa_loss} defines an equally weighted sum of task-normalized losses. The supervision is applied only to the target answer tokens, while the question tokens and continuous EEG tokens serve as conditional inputs.

\begin{figure*}[tbp]
    \centering
    \includegraphics[width=0.92\textwidth]{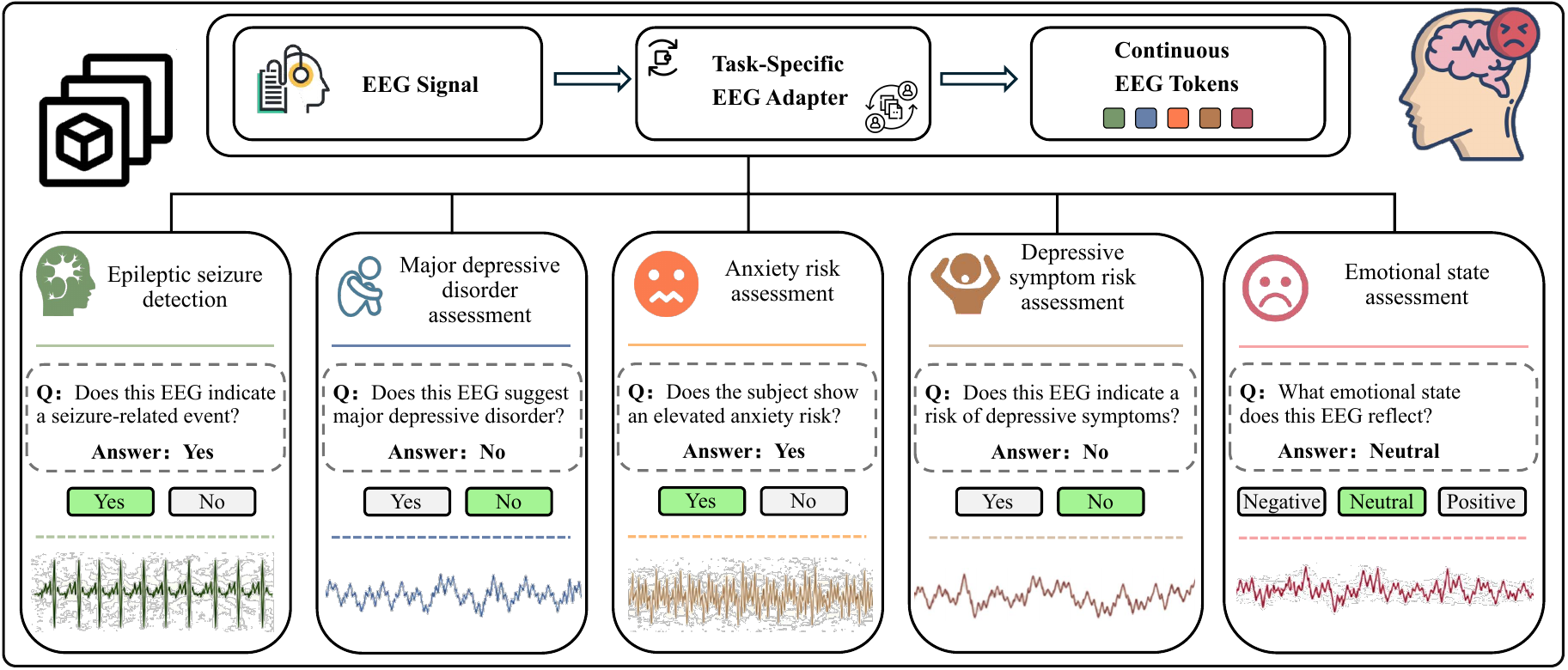}
    \caption{Closed-set generative QA formulation for the five EEG health assessment tasks in EEG-HealthBench.
    (a) Unified answer spaces across the five tasks.
    (b) Task-specific questions under a shared closed-set generative interface.}
    \label{fig:qa_formulation}
\end{figure*}

During inference, the model generates an output $\hat{y}_{r,i}$ and maps it to an answer in the corresponding predefined answer set $\mathcal{A}_r$. This formulation recasts EEG classification tasks with different label spaces as a unified conditional generation problem.
As summarized in Fig.~\ref{fig:qa_formulation}, the first four EEG health assessment tasks use binary \texttt{yes}/\texttt{no} answer spaces, whereas emotional state assessment uses the three-class answer space of negative, neutral, and positive. Although the task questions and answer spaces differ, all five tasks follow the same structured and parseable closed-set generative QA formulation.

\subsection{Overall Framework of \method{}}

\method{} extends the medical-image pathway of the underlying multimodal framework with a task-specific EEG pathway. As shown in Fig.~\ref{fig:framework}, the inherited medical-image pathway maps medical images into continuous image tokens, whereas the proposed EEG pathway maps heterogeneous EEG signals into continuous EEG tokens through task-specific EEG adapters. A unified QA template is processed by the Phi-3 tokenizer and text embedding layer to produce text tokens. Depending on the available non-textual evidence, image tokens or EEG tokens are combined with the corresponding text tokens to form an image--text or EEG--text input sequence. These are alternative modality-conditioned input sequences rather than simultaneous image--EEG inputs. The present study focuses on the EEG pathway and evaluates five heterogeneous EEG health assessment tasks; the image pathway is retained in Fig.~\ref{fig:framework} to clarify the architectural extension from the underlying multimodal framework.
The medical-image pathway is therefore shown only for architectural
completeness and does not participate in the training or evaluation
of the EEG experiments.

We initialize the language backbone from the publicly released Phi-3 checkpoint. All original Phi-3 parameters, including the text embedding layer, normalization layers, and LM head, remain frozen throughout EEG adaptation. During signal-side alignment, only the relevant EEG projectors are optimized; during language-side adaptation, only the shared LoRA adapter is optimized. For the $r$-th EEG health assessment task, the corresponding task-specific EEG adapter encodes the multichannel EEG signal and projects its representations into continuous EEG tokens in the Phi-3 input embedding space. These EEG tokens are combined with text tokens derived from the task-specific QA prompt to form an EEG--text input sequence, which is processed by Phi-3 with the shared language-side LoRA adapter to generate a closed-set health assessment answer.

This framework separates two modeling levels. On the signal side, each EEG task uses an independent EEG adapter to handle differences in channel configuration, temporal window, acquisition protocol, and signal distribution. On the language side, all EEG tasks share the same language-side LoRA adapter parameters to reuse representations associated with task questions and structured answer generation. This hierarchical design confines strongly task-specific EEG differences to low-level signal adaptation while maintaining high-level language-side sharing within the frozen Phi-3 backbone.

\subsection{Task-Specific EEG Signal-Side Adapter}

The goal of signal-side adaptation is to convert raw EEG inputs from different tasks into continuous EEG tokens that can be processed by Phi-3. Because EEG health assessment tasks differ in channel number, sampling rate, temporal window, and signal statistics, directly sharing a single EEG front-end can mix inconsistent spatio-temporal patterns and weaken task-relevant representations. We therefore assign an independent task-specific EEG adapter to each EEG health assessment task.

For the $r$-th task, the signal-side adaptation process is:
\begin{equation}
h^{\mathrm{eeg}}_{r,i}
=
E^{\mathrm{eeg}}_r
\left(
x^{\mathrm{eeg}}_{r,i}
\right),
\qquad
z^{\mathrm{eeg}}_{r,i}
=
P^{\mathrm{eeg}}_r
\left(
h^{\mathrm{eeg}}_{r,i}
\right),
\end{equation}
where $E^{\mathrm{eeg}}_r$ is the EEG encoder for task $r$, and
$P^{\mathrm{eeg}}_r$ denotes the task-specific projection pipeline. In the global-plus-sequence configuration, this pipeline comprises separate sequence and global projectors. The encoder extracts task-specific EEG representations, while the projection pipeline maps them into continuous EEG tokens. Together, they form the task-specific EEG adapter. We denote $z^{\mathrm{eeg}}_{r,i}\in\mathbb{R}^{L_r\times d}$, where $L_r$ is the number of continuous EEG tokens and $d$ is the hidden dimension of Phi-3.

For each EEG input, the task-specific EEG encoder produces two branch-wise token sequences. The two sequences are concatenated along the sequence dimension to form a sequence representation of shape $2N_r^{\mathrm{tok}}\times E$, where $N_r^{\mathrm{tok}}$ depends on the temporal length of task $r$, and $E=40$ is the EEG embedding dimension. Token-attention pooling is applied within the two branches, and the resulting pooled vectors are concatenated to form a global EEG representation of dimension $2E=80$. Separate multilayer perceptron (MLP) projectors, each implemented as a three-layer network, map the sequence representation and the global representation into the Phi-3 hidden dimension $d=3072$. The sequence projector produces $2N_r^{\mathrm{tok}}$ continuous EEG tokens, whereas the global projector maps the pooled global representation into one continuous EEG token.

In the sequence-only configuration, only the projected sequence tokens are used. In the global-plus-sequence configuration, the projected global token is prepended to the projected sequence tokens, resulting in $1+2N_r^{\mathrm{tok}}$ continuous EEG tokens. The global-plus-sequence configuration produces 63 continuous EEG tokens for Mumtaz, GAD-7, PHQ-9, and SEED, and 187 continuous EEG tokens for the epileptic seizure detection task. The resulting block of continuous EEG tokens is incorporated into the task-specific prompt at the designated modality position. Loss masking is applied so that the LM loss is computed only over the target answer tokens.

The EEG adapters are parameterized independently across tasks, which preserves task-specific signal structures and statistics. At the same time, all projectors map their outputs into the same hidden dimension as Phi-3. Therefore, heterogeneous raw EEG inputs are converted into a common token format and can be combined with the task question $q_r$ for closed-set answer generation.

\subsection{Shared Language-Side LoRA Adapter}

After task-specific EEG adaptation, the $i$-th sample from task $r$ is represented as continuous EEG tokens
$z^{\mathrm{eeg}}_{r,i}\in\mathbb{R}^{L_r\times d}$.
Let $t_r$ denote the text-token representation of the task-specific QA prompt. The continuous EEG tokens are incorporated into the prompt at the designated modality position to construct the Phi-3 input sequence:
\begin{equation}
H^{(0)}_{r,i}
=
\operatorname{Compose}
\left(
t_r,
z^{\mathrm{eeg}}_{r,i}
\right),
\end{equation}
where $\operatorname{Compose}(\cdot)$ preserves the token order of the prompt and inserts the continuous EEG tokens at the predefined position. In this sequence, the task prompt specifies the assessment semantics, while the EEG tokens provide the corresponding non-textual evidence.

To share language-side representations across EEG health assessment tasks, we introduce a shared LoRA adapter into the frozen Phi-3 backbone. For all tasks $r\in\mathcal{R}$, the input sequences are processed by the same shared language-side LoRA adapter:
\begin{equation}
H^{(L)}_{r,i}
=
\operatorname{Phi3}
\left(
H^{(0)}_{r,i};
\Theta_{0},
\Theta_{\mathrm{LoRA}}
\right),
\end{equation}
where $\Theta_{0}$ denotes the frozen pretrained parameters of Phi-3,
$\Theta_{\mathrm{LoRA}}$ denotes all trainable parameters of the
shared language-side LoRA adapter, including the low-rank branches
and routing network, and $L$ is the number of Phi-3 layers. During
adaptation, the Phi-3 backbone, text embedding layer, and LM head
remain frozen, while only $\Theta_{\mathrm{LoRA}}$ is updated.

LoRA is applied to the QKV projection and attention output projection in each self-attention sublayer, as well as the gated up-projection and down-projection in each feed-forward sublayer. For a target linear transformation $p$ in the $\ell$-th Phi-3 layer, standard LoRA~\cite{hu2021lora} is formulated as
\begin{equation}
F^{(\ell)}_{p}(U)
=
W^{(\ell)}_{0,p}U
+
s^{(\ell)}_{p}
B^{(\ell)}_{p}
A^{(\ell)}_{p}U,
\end{equation}
where $U$ denotes the input hidden representation,
$W^{(\ell)}_{0,p}$ is the frozen pretrained weight,
$A^{(\ell)}_{p}$ and $B^{(\ell)}_{p}$ are trainable low-rank matrices, and
$s^{(\ell)}_{p}$ is the scaling factor.

To accommodate variations in task semantics and input distributions, we employ a routed multi-branch LoRA implementation:
\begin{equation}
F^{(\ell)}_{p}(U)
=
W^{(\ell)}_{0,p}U
+
\sum_{m=1}^{M}
\rho^{(\ell)}_{p,m}(U)
s^{(\ell)}_{p,m}
B^{(\ell)}_{p,m}
A^{(\ell)}_{p,m}U,
\end{equation}
where $M$ is the number of low-rank branches,
$\left(A^{(\ell)}_{p,m},B^{(\ell)}_{p,m}\right)$ denotes the $m$-th branch, and
$\rho^{(\ell)}_{p,m}(U)$ is its input-dependent routing weight. This mechanism dynamically combines multiple low-rank branches according to the current hidden representation. The routed multi-branch implementation follows the efficient expert aggregation design used in HealthGPT~\cite{lin2025healthgpt}, while a single shared language-side LoRA adapter is used across all EEG health assessment tasks in our framework.

The final hidden representations are transformed into vocabulary
logits by the frozen LM head and autoregressively decoded according
to the conditional distribution defined in
Equation~\eqref{eq:qa_loss}. The generated output is then mapped to
the predefined answer set $\mathcal{A}_r$ of task $r$. In this manner,
task-specific EEG adapters handle low-level signal heterogeneity,
whereas the shared language-side LoRA adapter models high-level task
semantics and structured answer generation across tasks.

\subsection{Training Strategy}

Training includes signal-side alignment and unified language-side adaptation. For each task, the EEG adapter consists of a pretrained task-specific EEG encoder and corresponding EEG projectors. Each encoder is pretrained on its task training data and then frozen. During sequence-only alignment, only the sequence EEG projector is optimized; during global-plus-sequence alignment, both the sequence and global projectors are optimized while the encoder remains frozen. During single-task and unified language-side adaptation, all EEG encoders and projectors remain frozen.

Because EEG tasks differ in channel number, sampling rate, temporal window, and signal distribution, each task uses its own EEG adapter. These alignment stages provide signal-side representations for subsequent language-side adaptation.

The subsequent stage loads the aligned EEG adapters, freezes them and the original Phi-3 parameters, and updates only the language-side LoRA parameters. In single-task adaptation, an independent LoRA adapter is trained for each task, whereas \method{} shares one language-side LoRA adapter across all five tasks.

During unified adaptation (Stage 3-U), task-homogeneous mini-batches are used. Each mini-batch contains instances from one EEG health assessment task and activates only its task-specific EEG adapter, while the shared LoRA parameters are updated across tasks. At each update, one of the five tasks is sampled uniformly, followed by a mini-batch from its balanced training pool. Oversampling is applied only within the SEED training pool and does not alter the task-level sampling probability. The per-step unified language-side adaptation objective is

{\small
\begin{equation}
\mathcal{L}_{\mathrm{shared}}
=
-
\frac{1}{|\mathcal{B}|}
\sum_{(r,i)\in\mathcal{B}}
\frac{1}{T_i}
\sum_{t=1}^{T_i}
\log p_{\theta}
\bigl(
y_{r,i,t}
\mid
y_{r,i,<t}, q_r, z^{\mathrm{eeg}}_{r,i}
\bigr).
\label{eq:shared_loss}
\end{equation}
}

Here, $\mathcal{B}$ denotes the current mini-batch. Since all instances come from the same task, each update uses only the corresponding EEG adapter while updating the shared LoRA parameters. Equation~\eqref{eq:shared_loss} is the per-step loss during unified adaptation. Under the task-balanced homogeneous-batch protocol, the stochastic objective corresponds, in expectation, to an equally weighted average of task-normalized losses, equivalent to Equation~\eqref{eq:qa_loss} up to the constant factor $1/|\mathcal{R}|$, rather than a sample-frequency-weighted objective over raw dataset sizes. Thus, tasks with more raw recordings do not dominate the shared adapter through sample frequency.

\paragraph{Ethics Statement}
This study uses only existing publicly available EEG datasets and does
not involve new participant recruitment or data collection. Ethical
approval and informed consent for the original data collection were
handled by the respective dataset providers according to the protocols
reported for the source datasets.

\section{Experiments}

\subsection{EEG-HealthBench Benchmark and Evaluation Protocol}

To evaluate unified adaptation for heterogeneous cross-subject EEG health assessment, we establish EEG-HealthBench, a benchmark suite comprising five heterogeneous EEG health assessment paradigms constructed from five existing EEG resources under a unified subject-independent evaluation protocol and a closed-set generative QA format. EEG-HealthBench is not a newly collected single raw EEG dataset. Instead, it reorganizes multiple EEG resources under a common benchmark. The five paradigms cover epileptic seizure detection, major depressive disorder assessment, anxiety risk assessment, depressive symptom risk assessment, and emotional state assessment. The first four paradigms are organized as binary yes/no health assessment questions, whereas the emotional state paradigm is organized as a three-class negative/neutral/positive closed-set QA task. Each question-answer instance consists of an EEG signal, a task question, and a predefined answer. The model generates a closed-set answer conditioned on continuous EEG tokens and the task question.

To avoid subject information leakage, each task uses a fixed subject-independent test protocol. Test subjects are not involved in training, parameter updates, or model selection. All baseline methods and \method{} are compared under the same test protocol. The main text reports the benchmark organization and unified evaluation protocol. The task construction scripts, question templates, preprocessing procedures, and fixed splits are available in the public code
repository for reproducibility.

The five paradigms are constructed from existing public EEG resources.
Epileptic seizure detection is based on the CHB-MIT scalp EEG
database~\cite{chbmit2010}, while major depressive disorder assessment
uses the Mumtaz EEG dataset~\cite{mumtaz2017wavelet}. Anxiety risk and
depressive symptom risk assessment are constructed from the wireless
EEG datasets associated with the GAD-7 and PHQ-9 screening instruments,
respectively~\cite{sakib2025gad7eeg,spitzer2006gad7,
sakib2025phq9eeg,kroenke2001phq9}. Emotional state assessment is based
on SEED~\cite{zheng2015seed}.

The five resources differ substantially in cohort size, channel
configuration, sampling rate, temporal resolution, and model input
form. Table~\ref{tab:eeg_healthbench_tasks} summarizes the fixed
subject-independent partitions and signal characteristics used in
EEG-HealthBench.
\begin{table*}[tbp]
    \centering
    \caption{Cohort and signal characteristics of the five paradigms in EEG-HealthBench.}
\label{tab:eeg_healthbench_tasks}
    \tablefont
    \setlength{\tabcolsep}{3.5pt}
    \begin{tabular*}{\textwidth}{@{\extracolsep{\fill}}llcccc@{}}
        \toprule
        Paradigm
        & EEG resource
        & Cohort
        & Split
        & Channels / $f_s$
        & Answer space \\
        \midrule
        Epilepsy~\cite{chbmit2010}
        & CHB-MIT
        & 23 subjects / 24 cases
        & 15/4/5 cases
        & 21 / 256 Hz
        & yes / no \\
Mumtaz~\cite{mumtaz2017wavelet}
        & Mumtaz EEG
        & 63 subjects
        & 38/13/12 subjects
        & 19 / 256 Hz
        & yes / no \\
        GAD-7~\cite{sakib2025gad7eeg}
        & GAD-7 EEG
        & 38 subjects
        & 23/7/8 subjects
        & 14 / 128 Hz
        & yes / no \\
        PHQ-9~\cite{sakib2025phq9eeg}
        & PHQ-9 EEG
        & 31 subjects
        & 19/6/6 subjects
        & 14 / 128 Hz
        & yes / no \\
        SEED~\cite{zheng2015seed}
        & SEED
        & 15 subjects
        & 9/3/3 subjects
        & 62 / 200 Hz
        & negative / neutral / positive \\
        \bottomrule
    \end{tabular*}
    \vspace{2pt}
\end{table*}
\setlength{\tabcolsep}{4pt}
Benchmark unification preserves task-specific EEG characteristics while
standardizing the evaluation protocol and closed-set QA interface. During
Stage~3-U, the five heterogeneous tasks are sampled uniformly, ensuring
equal task-level participation despite differences in cohort size and
signal configuration.

Because tasks may exhibit class imbalance, balanced accuracy (BAC) is
used as the primary metric. Accuracy (ACC), macro-F1, and Cohen's
$\kappa$ are also reported to assess overall accuracy, class-balanced
performance, and chance-corrected agreement, respectively.

For a task $r$ with $K_r$ classes, BAC is defined as the average
class-wise recall:
\begin{equation}
\mathrm{BAC}_r
=
\frac{1}{K_r}
\sum_{k=1}^{K_r}
\frac{TP_k}{TP_k+FN_k},
\end{equation}
where $TP_k$ and $FN_k$ denote the numbers of true positives and false
negatives for class $k$, respectively.

ACC is defined as
\begin{equation}
\mathrm{ACC}_r
=
\frac{\sum_{k=1}^{K_r} TP_k}{N_r^{\mathrm{test}}},
\end{equation}
where $N_r^{\mathrm{test}}$ denotes the number of test instances in
task $r$.
Macro-F1 evaluates class-balanced performance by assigning equal weight
to the F1 score of each class:
\begin{equation}
\mathrm{MacroF1}_r
=
\frac{1}{K_r}
\sum_{k=1}^{K_r}
\frac{2TP_k}
{2TP_k+FP_k+FN_k},
\end{equation}
where $FP_k$ denotes the number of false positives for class $k$.
Cohen's $\kappa$ measures the agreement between predictions and
ground-truth labels after accounting for chance agreement:
\begin{equation}
\kappa_r
=
\frac{p_o-p_e}{1-p_e},
\end{equation}
where $p_o$ denotes the observed agreement and $p_e$ denotes the
agreement expected by chance from the empirical prediction and
ground-truth label marginals. Macro-F1 and Cohen's $\kappa$ are
computed separately for each task and then averaged equally across the
five tasks.

We also examine whether the generated outputs can be parsed into the
predefined answer spaces. Specifically, outputs from the first four
binary tasks are mapped to \texttt{yes} or \texttt{no}, whereas outputs
from SEED are mapped to negative, neutral, or positive. All generated
outputs in the five tasks can be mapped to their corresponding
predefined answer spaces, with no invalid outputs or abstentions.
Therefore, all evaluation metrics are computed from the parsed
closed-set answers. This setting unifies the output interfaces across
EEG health assessment tasks and does not imply open-ended medical
reasoning capability.

\subsection{Implementation Details}

Under the above benchmark and evaluation protocol, we divide signal-side alignment and language-side adaptation into several experimental stages. For each EEG health assessment task, we first construct a task-specific EEG adapter. The adapter consists of a task-specific EEG encoder, a sequence EEG projector, and a global EEG projector. The two projectors map sequence-level and global EEG representations, respectively, into continuous EEG tokens with the same hidden dimension as Phi-3. All single-task and unified adaptation experiments are based on this EEG-side representation.

\begin{figure}[htbp]
    \centering
    \includegraphics[width=\columnwidth]{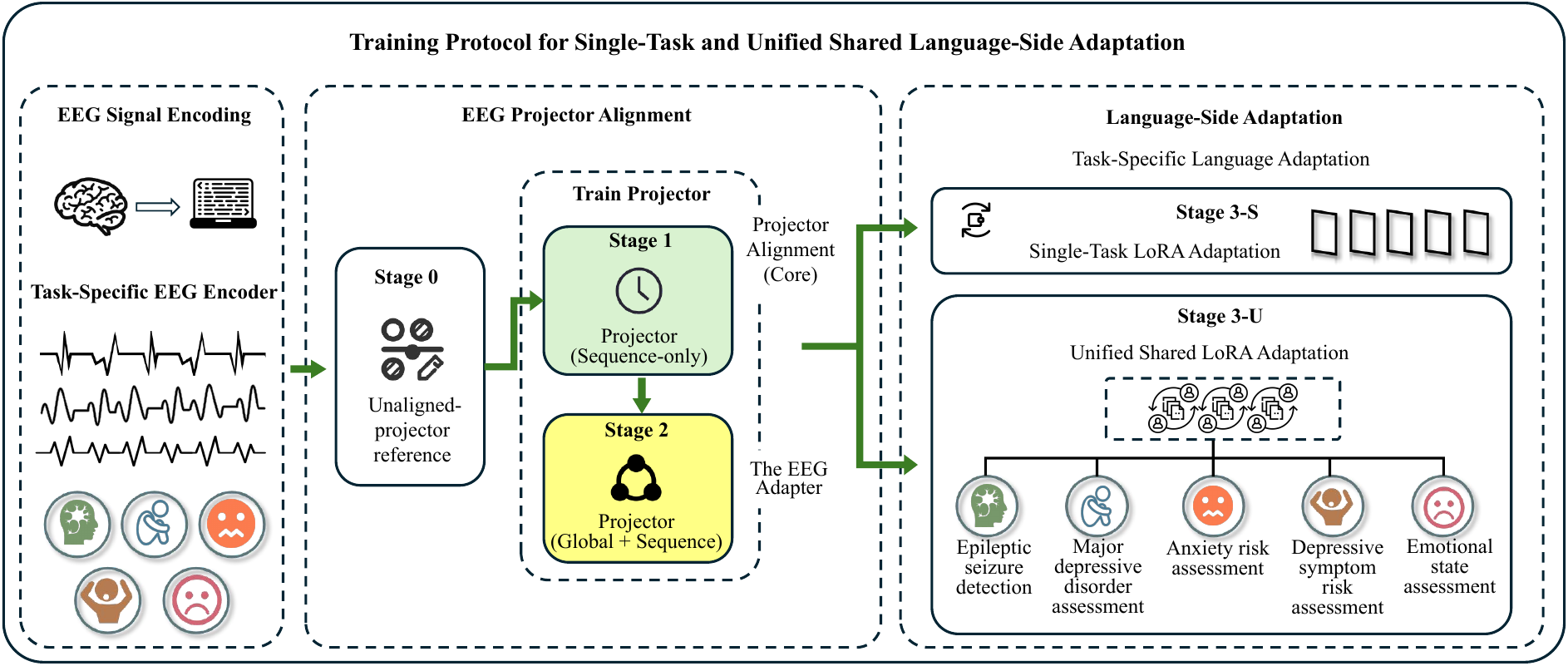}
    \caption{Training protocol for single-task and unified shared language-side adaptation.}
    \label{fig:training_protocol}
\end{figure}

The protocol is shown in Fig.~\ref{fig:training_protocol}, and the training objective and trainable modules of each stage are summarized in Table~\ref{tab:stage_protocol}. Stage~0 denotes an unaligned-projector reference evaluated before EEG projector alignment. In this setting, the EEG projectors remain at their initialization and are not optimized. This setting is included only to characterize performance before EEG--language projector alignment and should not be interpreted as conventional zero-shot EEG classification. Stage 1 optimizes the sequence EEG projector using sequence-only EEG tokens. Stage 2 optimizes both the sequence and global EEG projectors using global-plus-sequence EEG tokens.
The Stage 2 EEG adapters are used as the signal-side modules for subsequent single-task and unified experiments.

For language-side adaptation, Stage 3-S corresponds to the single-task LoRA pipeline, where a task-specific LoRA adapter is trained separately for each EEG health assessment task. Stage 3-U corresponds to the unified setting of \method{}, where one shared language-side LoRA adapter is trained across the five tasks. Therefore, Stage 3-S and Stage 3-U are parallel language-side adaptation settings based on the same set of Stage 2 EEG adapters. Their main difference is whether the language-side LoRA adapter parameters are task-specific or shared across tasks.

\begin{table*}[tbp]
    \centering
    \caption{Stage definitions for EEG projector alignment and language-side adaptation.}
    \label{tab:stage_protocol}
    \tablefont
    \begin{tabular*}{\textwidth}{@{\extracolsep{\fill}}llll@{}}
        \toprule
        Stage & Training objective & Trainable module & Role \\
        \midrule
        Stage 0 & Unaligned-projector evaluation & None & Pre-alignment reference \\
        Stage 1 & Sequence-only projector alignment & Sequence EEG projector & Sequence EEG token alignment \\
        Stage 2 & Global-plus-sequence projector alignment & Sequence and global EEG projectors & EEG adapters for downstream settings \\
        Stage 3-S & Single-task LoRA adaptation & Task-specific LoRA & Single-task language-side baseline \\
        Stage 3-U & Unified shared LoRA adaptation & Shared LoRA & Unified adaptation in this paper \\
        \bottomrule
    \end{tabular*}
\end{table*}

\paragraph{Training Configuration}
We use Phi-3-mini-4k-instruct as the language backbone. During unified language-side adaptation, all original Phi-3 parameters, including the LM head, and all task-specific EEG adapters remain frozen, while only the shared routed multi-branch LoRA adapter is optimized. The LoRA rank $r$ and scaling factor $\alpha$ are both set to 16, with a dropout rate of 0.05. Consistent with the architecture described in the Method section, LoRA is applied to the QKV and attention output projections in the self-attention sublayers, as well as the gated up-projection and down-projection in the feed-forward sublayers.

The unified model is trained for two epochs using AdamW with an initial learning rate of $1\times10^{-5}$, cosine learning-rate decay, a warmup ratio of 0.03, and no weight decay. The per-GPU batch size is 64, and the maximum sequence length is 4096. Training uses bfloat16 precision with TF32 enabled and gradient checkpointing. Following the task-balanced homogeneous-batch protocol described above, the five tasks are sampled uniformly, and each mini-batch contains samples from only one task. The shared LoRA adapter is initialized from scratch rather than warm-started from any task-specific adapter. Model selection is based on the average validation BAC, with the test sets used only for final evaluation. Experiments are conducted on a server equipped with eight NVIDIA A800 GPUs.

The emotional state assessment task is included in unified adaptation to examine whether the framework can extend from epileptic seizure detection and mental health assessment to a broader EEG brain-state assessment scenario. In addition to \method{}, we compare representative convolutional, recurrent, Transformer-based, and EEG foundation models, including EEGNet~\cite{lawhern2018eegnet}, DeepConvNet and ShallowConvNet~\cite{schirrmeister2017deep}, LSTM~\cite{hochreiter1997lstm}, CNN--LSTM~\cite{chakravarthi2022cnnlstm}, Conformer~\cite{song2023eegconformer}, LMDA-Net~\cite{miao2023lmda}, BIOT~\cite{yang2023biot}, MGFormer~\cite{rahman2025mgformer}, MSCFormer~\cite{zhao2025mscformer}, TMSA-Net~\cite{zhao2025tmsanet}, LaBraM~\cite{jiang2024labram}, EEG-DINO~\cite{wang2025eegdino}, and LUNA~\cite{doner2025luna}. We further include NeuroLM-B~\cite{jiang2024neurolm} as a unified EEG--language multi-task baseline. The single-task LoRA pipeline, corresponding to Stage 3-S in Table~\ref{tab:stage_protocol}, is also evaluated to examine whether shared language-side adaptation provides benefits over independent task-specific adaptation. BAC is used as the primary metric across the five tasks, while ACC is reported in the stage-wise analysis as an auxiliary metric.

\subsection{Main Results on Five EEG Health Assessment Tasks}
We first compared the overall performance of different methods across the five EEG health assessment tasks in EEG-HealthBench. As shown in Table~\ref{tab:main_results}, \method{} achieved an average BAC of 69.58\%, an average accuracy of 69.50\%, an average macro-F1 of 68.75\%, and a mean Cohen's $\kappa$ of 0.425. It ranked first among all evaluated methods in both average BAC and macro-F1. Relative to EEG-DINO, the closest discriminative model, \method{} improved average BAC and macro-F1 by 0.05 and 0.71 percentage points, respectively, while EEG-DINO retained a slightly higher mean $\kappa$. These results indicate that \method{} achieves competitive task-averaged performance under equal task weighting and class-balanced evaluation.

Compared with the single-task LoRA pipeline, unified shared LoRA increased average BAC from 67.70\% to 69.58\% and average macro-F1 from 66.67\% to 68.75\%. These improvements show that sharing language-side adaptation across tasks improves overall class-balanced performance under the same unified question-answering interface. Under our task-balanced multi-task tuning protocol, NeuroLM-B achieved an average BAC of 67.84\%, an average accuracy of 70.13\%, an average macro-F1 of 66.79\%, and a mean Cohen's $\kappa$ of 0.412. Although NeuroLM-B achieved the highest average accuracy among the evaluated methods, \method{} outperformed it by 1.74 percentage points in average BAC and 1.96 percentage points in average macro-F1, indicating more balanced performance across tasks and classes.
This distinction is important for EEG-HealthBench because several tasks
exhibit class imbalance, making accuracy sensitive to majority-class
performance. In contrast, BAC weights class-wise recall equally and is
therefore better aligned with the benchmark objective of evaluating
balanced performance across heterogeneous assessment tasks. The corresponding macro-F1 results provide a complementary
class-balanced perspective and support the same overall conclusion.

The task-level results further highlight the heterogeneity of
EEG-HealthBench. EEG-HealthLLM achieves BAC values of 72.78\%,
93.91\%, 49.94\%, 70.62\%, and 60.67\% on Epilepsy, Mumtaz,
GAD-7, PHQ-9, and SEED, respectively. The substantial variation
across tasks indicates that unified adaptation does not eliminate
differences in task difficulty. Instead, the proposed framework
provides a shared language-side interface while retaining
task-specific signal pathways for heterogeneous EEG distributions.

\begin{table*}[tbp]
    \centering
    \caption{Overall task-averaged performance on EEG-HealthBench.}
    \label{tab:main_results}
    \tablefont
    \setlength{\tabcolsep}{3.5pt}
    \begin{tabular*}{\textwidth}{@{\extracolsep{\fill}}llccccc@{}}
        \toprule
        Model & Adaptation setting & Unified interface & Avg. BAC (\%) $\uparrow$ & Avg. Acc (\%) $\uparrow$ & Avg. macro-F1 (\%) $\uparrow$ & Mean $\kappa$ $\uparrow$ \\
        \midrule
        \multicolumn{7}{l}{\textit{Discriminative EEG models}} \\
        EEG-DINO~\cite{wang2025eegdino} & Task-specific & No & \underline{69.53} & \underline{70.09} & \underline{68.04} & \textbf{0.429} \\
        BIOT~\cite{yang2023biot} & Task-specific & No & 68.40 & 68.11 & 66.38 & 0.406 \\
        MGFormer~\cite{rahman2025mgformer} & Task-specific & No & 68.12 & 69.31 & 64.71 & 0.396 \\
        MSCFormer~\cite{zhao2025mscformer} & Task-specific & No & 66.98 & 68.22 & 65.50 & 0.373 \\
        EEGNet~\cite{lawhern2018eegnet} & Task-specific & No & 66.40 & 64.85 & 64.03 & 0.356 \\
        LMDA-Net~\cite{miao2023lmda} & Task-specific & No & 64.63 & 64.11 & 62.21 & 0.317 \\
        Conformer~\cite{song2023eegconformer} & Task-specific & No & 64.44 & 67.70 & 62.78 & 0.329 \\
        ShallowConvNet~\cite{schirrmeister2017deep} & Task-specific & No & 64.41 & 64.83 & 62.53 & 0.323 \\
        DeepConvNet~\cite{schirrmeister2017deep} & Task-specific & No & 63.52 & 63.55 & 61.44 & 0.328 \\
        LUNA~\cite{doner2025luna} & Task-specific & No & 63.27 & 66.46 & 62.18 & 0.317 \\
        TMSA-Net~\cite{zhao2025tmsanet} & Task-specific & No & 62.71 & 61.91 & 60.98 & 0.299 \\
        LaBraM~\cite{jiang2024labram} & Task-specific & No & 61.95 & 65.03 & 60.80 & 0.288 \\
        LSTM~\cite{hochreiter1997lstm} & Task-specific & No & 59.81 & 62.99 & 59.20 & 0.259 \\
        CNN-LSTM~\cite{chakravarthi2022cnnlstm} & Task-specific & No & 59.60 & 60.53 & 54.09 & 0.255 \\
        \midrule
        \multicolumn{7}{l}{\textit{Language-interface models}} \\
        NeuroLM-B~\cite{jiang2024neurolm} & Task-balanced multi-task tuning & Yes & 67.84 & \textbf{70.13} & 66.79 & 0.412 \\
        Single-task LoRA pipeline & Task-specific LoRA & Yes & 67.70 & 67.77 & 66.67 & 0.387 \\
        \method{} (Ours) & Unified shared LoRA & Yes & \textbf{69.58} & 69.50 & \textbf{68.75} & \underline{0.425} \\
        \bottomrule
    \end{tabular*}
\end{table*}
\setlength{\tabcolsep}{4pt}

\begin{figure*}[bp]
    \centering
    \includegraphics[width=\textwidth]{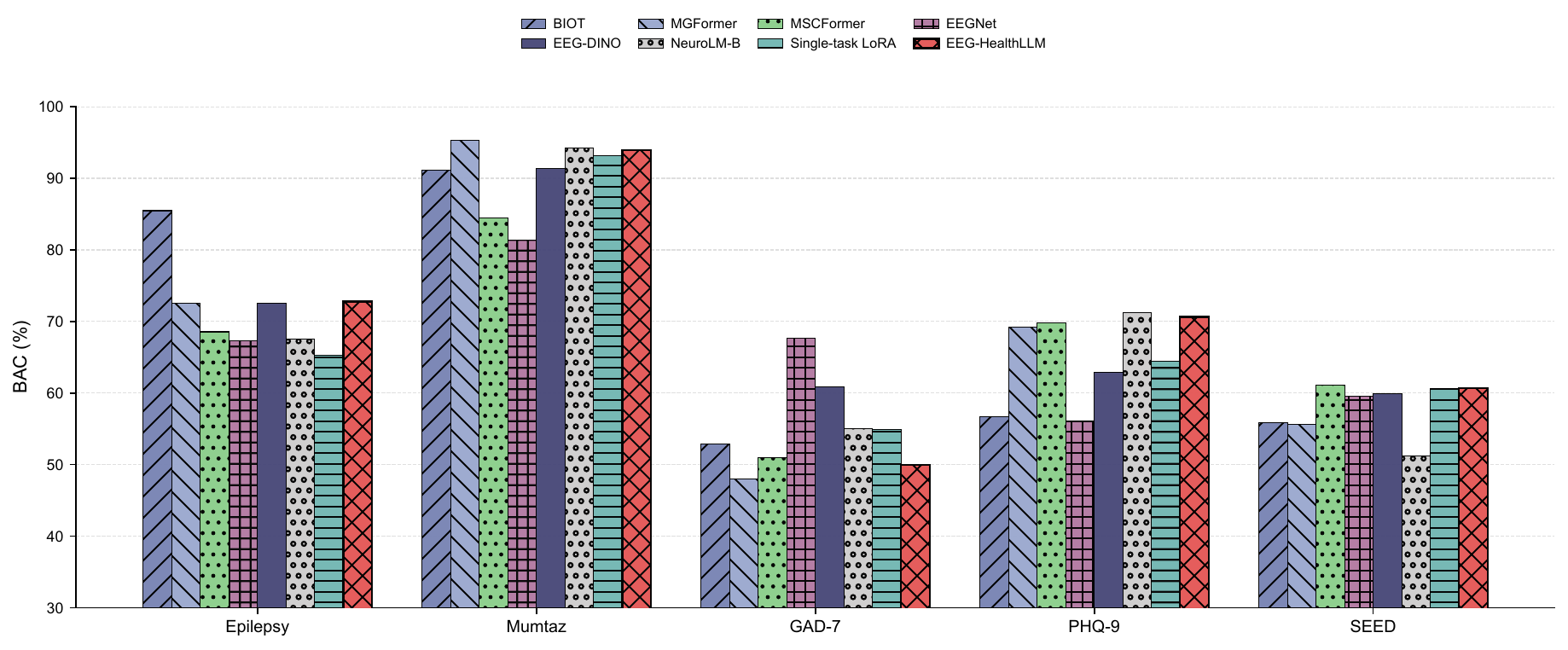}
    \caption{Task-level BAC comparison of representative methods on EEG-HealthBench.}
    \label{fig:main_results_bar}
\end{figure*}

\subsection{Effectiveness of Shared Language-Side Adaptation}

We further examine shared versus independent language-side adaptation. Table~\ref{tab:single_task_stages} reports BAC/ACC across Stages~0--3-S. Stage~3-S exceeds Stage~0 on all tasks, while the gains in Stages~1 and~2 vary across tasks. The near-chance Stage~0 performance indicates that the unaligned projectors provide little effective EEG evidence and is therefore included only as a pre-alignment reference. Table~\ref{tab:single_vs_unified} further compares task-specific and unified shared LoRA in terms of BAC and adapter parameter footprint.

Unified shared LoRA improves the average BAC from 67.70\% to 69.58\%,
with task-wise changes of $+7.46$, $+0.69$, $-4.95$, $+6.15$, and
$+0.08$ percentage points. GAD-7 is the only task that exhibits a
substantial decrease, indicating that the benefit of shared
language-side adaptation is task dependent. Because both settings use
the same Stage~2 EEG adapters and differ primarily in whether
language-side LoRA is shared, the average gain is consistent with a
contribution from shared language-side adaptation.

\begin{table}[tbp]
    \centering
    \caption{Stage-wise performance of the single-task LoRA pipeline (\%). Each cell reports BAC/ACC.}
    \label{tab:single_task_stages}
    \tablefont
    \begin{tabular*}{\columnwidth}{@{\extracolsep{\fill}}lcccc@{}}
        \toprule
        Task & Stage 0 & Stage 1 & Stage 2 & Stage 3-S \\
        \midrule
        Epilepsy & 50.00/50.00 & 51.41/51.41 & 64.52/64.52 & 65.32/65.32 \\
        Mumtaz & 50.00/52.98 & 90.28/90.51 & 91.66/91.95 & 93.22/93.31 \\
        GAD-7 & 50.00/37.50 & 50.33/37.13 & 50.00/37.50 & 54.89/57.92 \\
        PHQ-9 & 50.00/33.15 & 52.72/37.88 & 50.00/33.43 & 64.47/61.28 \\
        SEED & 32.27/31.85 & 60.01/60.42 & 60.21/60.54 & 60.59/61.02 \\
        \bottomrule
    \end{tabular*}
\end{table}

Each Stage~3-S or Stage~3-U run contains 27.39M trainable language-side adapter parameters. Supporting all five tasks with independent Stage~3-S adapters therefore requires 136.97M language-side parameters, whereas \method{} uses a single 27.39M shared adapter. With the same 19.08M task-specific EEG-adapter footprint in both settings, unified adaptation reduces the aggregate language-side footprint by 80.0\% and the total adapter footprint from 156.05M to 46.47M, corresponding to a 70.2\% reduction.

\begin{table*}[tbp]
    \centering
    \caption{Comparison between single-task and unified shared language-side adaptation. Task performance is reported as BAC (\%), and parameter footprints correspond to all five tasks.}
    \label{tab:single_vs_unified}
    \tablefont
    \begin{tabular*}{\textwidth}{@{\extracolsep{\fill}}lccccccccc@{}}
        \toprule
        Setting & Epilepsy & Mumtaz & GAD-7 & PHQ-9 & SEED & Avg. BAC
        & \shortstack{Language-side\\footprint}
        & \shortstack{EEG-adapter\\footprint}
        & \shortstack{Total adapter\\footprint} \\
        \midrule
        Five task-specific LoRA adapters (Stage 3-S)
        & 65.32 & 93.22 & 54.89 & 64.47 & 60.59 & 67.70
        & 136.97M & 19.08M & 156.05M \\
        Unified shared LoRA (Stage 3-U)
        & 72.78 & 93.91 & 49.94 & 70.62 & 60.67 & 69.58
        & 27.39M & 19.08M & 46.47M \\
        \bottomrule
    \end{tabular*}
\end{table*}

Taken together, these results show that task-specific signal-side
adaptation and shared language-side adaptation provide a practical
balance between task specialization, cross-task sharing, and
parameter efficiency.

\subsection{Task Trade-Off Analysis in Unified Adaptation}

After confirming the average benefit of shared language-side adaptation,
we further investigate the task-dependent behavior observed under
unified adaptation from task-expansion, signal-representation, and
optimization perspectives. From a broader multi-task optimization
perspective, jointly trained tasks may compete over a shared parameter
space, making task trade-offs a multi-objective optimization
problem~\cite{sener2018multitask}.

\begin{figure}[htbp]
    \centering
    \includegraphics[width=\columnwidth]{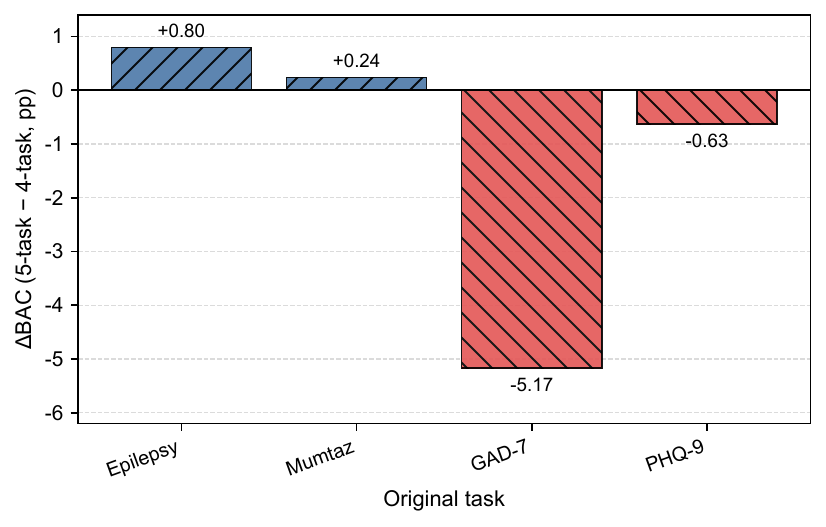}
    \caption{BAC changes of the original four tasks after extending unified adaptation from four to five tasks.}
    \label{fig:paradigm_expansion}
\end{figure}

\begin{figure}[htbp]
    \centering
    \includegraphics[width=\columnwidth]{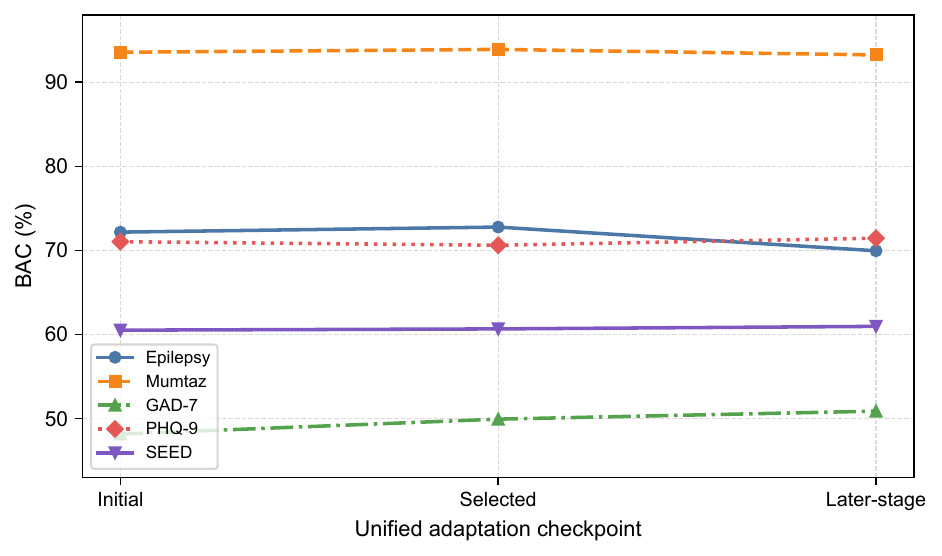}
    \caption{Task-level BAC at different checkpoints during five-task unified adaptation.}
    \label{fig:training_dynamics}
\end{figure}

Figure~\ref{fig:paradigm_expansion} compares the original four tasks before and after adding the emotional state assessment task. The four-task average BAC of 73.00\% and the five-task average BAC of 69.58\% should not be interpreted as a direct superiority comparison because the task sets differ in number and heterogeneity. After adding the emotional state assessment task, epileptic seizure detection increases from 71.98\% to 72.78\%, major depressive disorder assessment increases from 93.67\% to 93.91\%, and depressive symptom risk assessment decreases slightly from 71.25\% to 70.62\%. The main change is in anxiety risk assessment, whose BAC decreases from 55.11\% to 49.94\%. Thus, the observed decrease is concentrated on one task rather than shared by all original tasks. To examine whether this trade-off persists during training, Fig.~\ref{fig:training_dynamics} shows task-level performance at different checkpoints in the same unified adaptation process.

Figure~\ref{fig:training_dynamics} further shows that task-level
performance evolves differently during unified adaptation. Improvements
in GAD-7, PHQ-9, or SEED can coincide with decreases in epileptic
seizure detection, confirming that the shared adaptation process
involves task-dependent trade-offs rather than uniform gains.

\paragraph{Signal-Side Representation Analysis}

To further investigate why GAD-7 exhibits the largest degradation
under unified adaptation, we examine whether its EEG representations
already show unusually weak label-associated structure before shared
language-side adaptation. For the subject-level assessment tasks,
segment-level representations are first aggregated within each
subject.

We focus the main comparison on Mumtaz, GAD-7, and PHQ-9 because
they share a subject-level assessment setting, whereas seizure
detection and SEED use different supervision units. In particular,
GAD-7 and PHQ-9 have the same channel configuration and sampling
rate, making their contrast especially informative given their
opposite performance changes under unified adaptation.

We evaluate representation separability using three complementary
statistics. The class-balanced Fisher ratio measures the separation
between class centers relative to within-class dispersion, with larger
values indicating clearer class structure. Macro silhouette evaluates
whether samples are closer to representations from the same class than
to those from other classes while assigning equal weight to each class.
The permutation percentile measures how highly the observed Fisher
ratio ranks among all class-count-preserving label permutations,
thereby assessing whether the observed label-associated geometry is
stronger than that expected from arbitrary label assignments.

\begin{table}[tbp]
    \centering
    \caption{Signal-side representation separability before shared language-side adaptation.}
    \label{tab:representation_separability}
    \tablefont
    \begin{tabular*}{\columnwidth}{@{\extracolsep{\fill}}lccc@{}}
        \toprule
        Task & Fisher ratio & Macro Silhouette & Permutation percentile \\
        \midrule
        Mumtaz & 2.81 & 0.60 & 100.0\% \\
        GAD-7  & 1.01 & 0.30 & 98.2\% \\
        PHQ-9  & 1.15 & 0.27 & 93.3\% \\
        \bottomrule
    \end{tabular*}
\end{table}

As shown in Table~\ref{tab:representation_separability}, GAD-7 and
PHQ-9 exhibit comparable signal-side class geometry despite their
opposite performance changes under unified adaptation. Their Fisher
ratios are 1.01 and 1.15, respectively, while their macro silhouette
scores are 0.30 and 0.27. Moreover, the observed GAD-7 Fisher ratio
lies at the 98.2nd percentile of all class-count-preserving label
permutations, indicating non-random label-associated structure in its
frozen EEG representation. Mumtaz exhibits substantially stronger
subject-level separability, with a Fisher ratio of 2.81 and a macro
silhouette score of 0.60.

We further compare the representations before and after Stage~2
projection and observe no degradation of the GAD-7 class structure
after projection. These results provide no evidence that the GAD-7
performance drop is primarily caused by uniquely weak signal-to-label
separability or by loss of class structure during EEG--language
alignment. We therefore next examine whether the observed behavior is
associated with optimization-level interactions among tasks.

\paragraph{Gradient Conflict Analysis}

To examine optimization-level interactions among tasks, we compute
gradient cosine similarity with respect to the trainable shared LoRA
parameters~\cite{yu2020gradient,sun2026taskpreferred}. Let $\phi$
denote the shared LoRA parameters and
$g_{r,b}=\nabla_{\phi}\mathcal{L}_{r,b}$ the gradient of the $b$-th
mini-batch from task $r$. The empirical task-level gradient direction
is defined as
\begin{equation}
\bar{g}_r =
\frac{1}{B_r}
\sum_{b=1}^{B_r}
\frac{g_{r,b}}
{\|g_{r,b}\|_2},
\end{equation}
where $B_r$ is the number of sampled mini-batches. The cosine
similarity between tasks $r$ and $s$ is
\begin{equation}
\mathrm{sim}(r,s)
=
\frac{\bar g_r^{\top}\bar g_s}
{\|\bar g_r\|_2\|\bar g_s\|_2},
\end{equation}
where negative values indicate conflicting optimization directions.
We further summarize the average negative conflict of task $r$ as
\begin{equation}
\mathrm{Conflict}(r)
=
\frac{1}{|\mathcal{R}|-1}
\sum_{s\neq r}
\max\left(0,-\mathrm{sim}(r,s)\right).
\end{equation}

The analysis uses the unified shared LoRA checkpoint from the main
experiment. An equal number of mini-batches is sampled from each task
under the Stage~3-U protocol, and gradients are computed using the
teacher-forced generation loss.

\begin{table*}[tbp]
    \centering
    \caption{Task-specific residual LoRA ablation results. All values are BAC (\%).}
    \label{tab:sr_lora_ablation}
    \tablefont
    \begin{tabular*}{\textwidth}{@{\extracolsep{\fill}}lcccccc@{}}
        \toprule
        Setting & Epilepsy & Mumtaz & GAD-7 & PHQ-9 & SEED & Avg. \\
        \midrule
        Single-task LoRA & 65.32 & 93.22 & \textbf{54.89} & 64.47 & 60.59 & 67.70 \\
        Unified shared LoRA & \textbf{72.78} & \textbf{93.91} & 49.94 & 70.62 & 60.67 & \textbf{69.58} \\
        SR-LoRA-r2 & 70.36 & 93.09 & 47.50 & \textbf{71.46} & \textbf{60.73} & 68.63 \\
        SR-LoRA-r4 & 70.97 & 93.01 & 48.17 & 70.62 & 60.61 & 68.68 \\
        \bottomrule
    \end{tabular*}
\end{table*}

\begin{figure}[htbp]
    \centering
\includegraphics[width=\columnwidth]{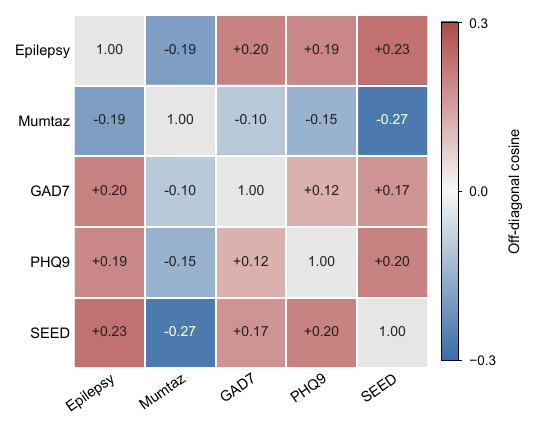}
    \caption{Gradient cosine similarity across the five EEG health assessment tasks on shared LoRA parameters. Positive and negative values indicate aligned and conflicting gradient directions, respectively.}
    \label{fig:gradient_conflict}
\end{figure}

In Figure~\ref{fig:gradient_conflict}, we reveal several conflicting
cross-task gradients. The strongest negative similarity occurs between
Mumtaz and SEED ($-0.27$), with additional conflicts for
Epilepsy--Mumtaz ($-0.19$), Mumtaz--PHQ-9 ($-0.15$), and
Mumtaz--GAD-7 ($-0.10$); all other task pairs are positively aligned.
Thus, although unified shared LoRA improves average performance, the
shared optimization remains task dependent, consistent with the
trade-offs observed during training.
Notably, the degree of gradient conflict does not simply correspond to
test performance. Mumtaz has the highest average conflict score, but
its test BAC remains 93.91\%. In contrast, GAD-7 has a relatively low
conflict score but shows the largest degradation under unified
adaptation. Together with the preceding signal-side representation
analysis, these results indicate that the GAD-7 decrease cannot be
explained solely by either unusually weak pre-sharing class
separability or gradient conflict among shared LoRA updates. The
observed behavior is therefore task dependent and may involve more
complex interactions during heterogeneous multi-task adaptation.

\paragraph{Task-Specific Residual LoRA Analysis}
In addition to gradient analysis, we examine whether task-specific
low-rank compensation can alleviate the observed trade-off by adding
a residual LoRA to the trained shared LoRA. The Phi-3 backbone,
task-specific EEG adapters, and shared LoRA remain frozen, and only
the residual LoRA associated with the current task is optimized under
the same task-homogeneous training protocol. We denote residual LoRA
ranks of 2 and 4 as SR-LoRA-r2 and SR-LoRA-r4, respectively.
As shown in Table~\ref{tab:sr_lora_ablation}, SR-LoRA-r2 and
SR-LoRA-r4 achieve average BAC values of 68.63\% and 68.68\%,
respectively, both below unified shared LoRA at 69.58\%. Their GAD-7
performance also remains lower than the shared baseline. These results
indicate that simply adding task-specific residual capacity does not
resolve the observed trade-off under the evaluated setting.

Taken together, the representation, gradient, and residual-adaptation
analyses suggest that the GAD-7 degradation cannot be attributed to a
single factor. In particular, we find no evidence that it is primarily
caused by uniquely weak pre-sharing class separability, excessive
gradient conflict, or insufficient task-specific language-side
capacity. The observed trade-off therefore appears to reflect more
complex task-dependent interactions during heterogeneous multi-task
adaptation, motivating more adaptive cross-task coordination in future
work.

\section{Conclusion}

This paper proposes EEG-HealthLLM, a Phi-3-based adaptation framework
for heterogeneous cross-subject EEG health assessment. The framework
employs task-specific EEG adapters to preserve task-dependent signal
characteristics and shares language-side adaptation parameters across
five closed-set generative question answering tasks, thereby balancing
low-level signal heterogeneity with high-level language modeling
consistency. Experimental results on EEG-HealthBench show that the
proposed method achieves competitive average performance under
subject-independent evaluation while reducing the aggregate
language-side adapter footprint by 80.0\%. Rather than replacing
conventional task-specific classifiers, EEG-HealthLLM provides a
unified and parameter-efficient language interface for heterogeneous
EEG health assessment. Future work will explore more adaptive
cross-task coordination, cross-center generalization, and open-ended
medical explanation.

\bibliographystyle{IEEEtran}
\bibliography{uni_healthgpt_refs}

\end{document}